\documentclass{article}

\usepackage{arxiv}

\usepackage{amsmath,amsfonts,bm}

\def\eqref#1{equation~\ref{#1}}

\def\1{\bm{1}}

\DeclareMathAlphabet{\mathsfit}{\encodingdefault}{\sfdefault}{m}{sl}
\SetMathAlphabet{\mathsfit}{bold}{\encodingdefault}{\sfdefault}{bx}{n}

\usepackage[utf8]{inputenc} 
\usepackage[T1]{fontenc}    
\usepackage{booktabs}       
\usepackage{amsfonts}       
\usepackage{nicefrac}       
\usepackage{microtype}      
\usepackage{lipsum}         
\usepackage{natbib}
\usepackage{doi}
\usepackage{wrapfig}
\usepackage{hyperref}

\usepackage{cleveref}       
\usepackage{url}
\usepackage{graphicx}
\usepackage{multirow}
\usepackage{colortbl}
\usepackage{xcolor}
\usepackage{tabularx}
\usepackage{longtable}
\usepackage{caption}
\newcolumntype{C}{>{\centering\arraybackslash}X}
\newcolumntype{g}{>{\columncolor{blue!10}}c}

\title{Beyond Gradient and Priors in Privacy Attacks: Leveraging Pooler Layer Inputs of Language Models in Federated Learning\thanks{
Our code is available at \href{https://github.com/mexiQQ/privacy_attack_llm}{https://github.com/mexiQQ/privacy\_attack\_llm}}}

\newif\ifuniqueAffiliation
\uniqueAffiliationtrue

\ifuniqueAffiliation 
\author{ Jianwei Li\thanks{Department of Computer Science at North Carolina State University, Email: \texttt{jli265@ncsu.edu}} \\
	\and
        Sheng Liu\thanks{Center for Data Science at New York University; Stanford University, Email: \texttt{shengl@stanford.edu}} \\
	\and
	  Qi Lei\thanks{Center for Data Science at New York University, Email: \texttt{ql518@nyu.edu}}\\ 
}
\else
\usepackage{authblk}

\newbox{\orcid}\sbox{\orcid}{\includegraphics[scale=0.06]{orcid.pdf}} 
\author[1]{%
	\href{https://orcid.org/0000-0000-0000-0000}{\usebox{\orcid}\hspace{1mm}David S.~Hippocampus\thanks{\texttt{hippo@cs.cranberry-lemon.edu}}}%
}
\author[1,2]{%
	\href{https://orcid.org/0000-0000-0000-0000}{\usebox{\orcid}\hspace{1mm}Elias D.~Striatum\thanks{\texttt{stariate@ee.mount-sheikh.edu}}}%
}
\affil[1]{Department of Computer Science, Cranberry-Lemon University, Pittsburgh, PA 15213}
\affil[2]{Department of Electrical Engineering, Mount-Sheikh University, Santa Narimana, Levand}
\fi

\renewcommand{\shorttitle}{}

\hypersetup{
pdftitle={A template for the arxiv style},
pdfsubject={q-bio.NC, q-bio.QM},
pdfauthor={David S.~Hippocampus, Elias D.~Striatum},
pdfkeywords={First keyword, Second keyword, More},
}

\begin{document}
\maketitle

\begin{abstract}

Language models trained via federated learning (FL) demonstrate impressive capabilities in handling complex tasks while protecting user privacy. Recent studies indicate that leveraging gradient information and prior knowledge can potentially reveal training samples within FL setting. However, these investigations have overlooked the potential privacy risks tied to the intrinsic architecture of the models. This paper presents a two-stage privacy attack strategy that targets the vulnerabilities in the architecture of contemporary language models, significantly enhancing attack performance by initially recovering certain feature directions as additional supervisory signals. Our comparative experiments demonstrate superior attack performance across various datasets and scenarios, highlighting the privacy leakage risk associated with the increasingly complex architectures of language models. We call for the community to recognize and address these potential privacy risks in designing large language models.

\end{abstract}

\newpage

\newcounter{myCounter}
\section{Introduction}

Language models trained under the Federated Learning paradigm play a pivotal role in diverse applications such as next-word predictions on mobile devices and electronic health record analysis in hospitals~\citep{ramaswamy2019federated, li2020federated}. This training paradigm prioritizes user privacy by restricting raw data access to local devices and centralizing only the model's updates, such as gradients and parameters~\citep{mcmahan2017communication}. While the FL framework is created to protect user privacy, vulnerabilities still persist. Previous works~\citep{geiping2020inverting, yin2021see, jeon2021gradient} proved that attackers can almost perfectly recover image data with gradients, which also highlighted the potential risks in the realm of textual data.

Many studies have investigated vulnerabilities of private data in FL when applied to language models~\citep{zhu2019deep, deng2021tag, balunovic2022lamp, gupta2022recovering}. \citet{zhu2019deep} and \citet{deng2021tag} leverage gradient information and well-designed objective functions to build an optimization-based pipeline that can recover certain training textual data in minimal batch sizes. \citet{balunovic2022lamp} and \citet{gupta2022recovering} further improve the recovery rate by incorporating prior knowledge embedded in LLM to provide additional optimization or retrieval signals. All of them mainly focus on algorithm design and external information but rarely notice inherent privacy leakage risks embedded in the language models themselves. In contrast to these approaches, ~\citet{fowl2022decepticons} and ~\citet{boenisch2023curious} assume a malicious central server, designing specific malicious parameters and architectures for language models to enhance attack performance. To a certain degree, their research is connected to the privacy vulnerabilities of specific module designs. However, their methods depend on matched tampered parameters, such as identity weight matrices. These approaches also violate the training objective, generating useless gradients and parameter updates. 

    

\begin{wrapfigure}{r}{0.5\textwidth}
  \centering
  \includegraphics[width=0.48\textwidth]{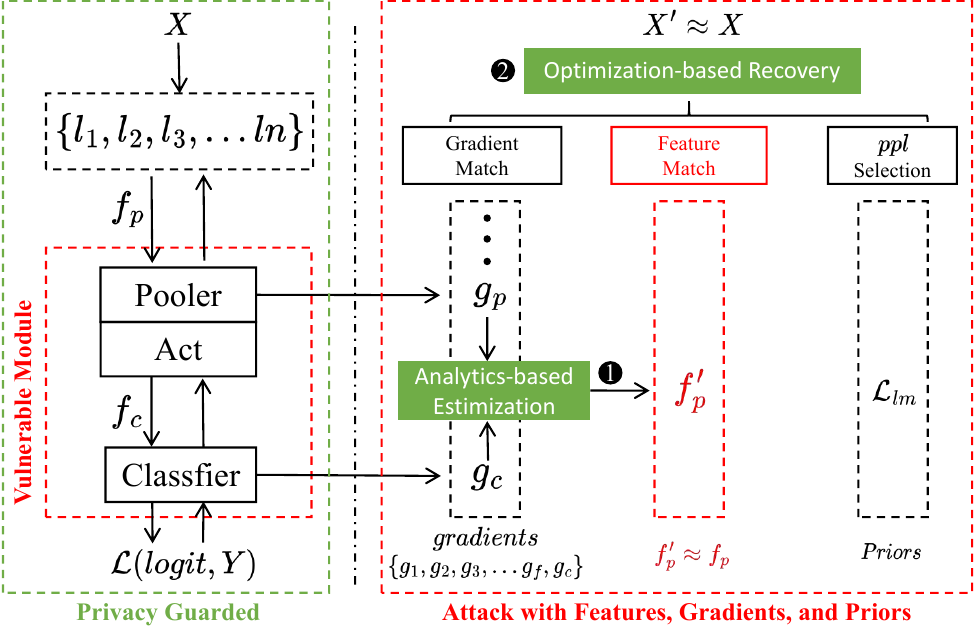}
  \captionof{figure}{Schematic illustration of the two-stage privacy attack method proposed in this study.~\textbf{1})~The first stage involves the analytics-based reconstruction of the feature information associated with the specific Pooler layer in Transformer-based language models.~\textbf{2})~The second stage utilizes the reconstructed feature information, combined with gradient inversion and prior knowledge, to guide the recovery of training data. This figure highlights the approach of intermediate feature recovery while exposing the inherent privacy risks in contemporary language model architectures. }
  \label{fig:3}
  \vspace{-0.3cm}
\end{wrapfigure}

Recently, \citet{wang2023reconstructing} provides a theoretical analysis that recovers training samples from gradient information for a two-layer fully connected network. A major limitation of their approach is the reliance on a randomly initialized network instead of an actual, pre-trained network. This reliance poses challenges in applying their insights to the prevalent training paradigm that fine-tuning pre-trained models. When applied to deeper networks, their method also relies on identity modules and other transparently detectable weight manipulations, limiting its practical utility. However, building upon this foundation, we have found that certain prevalent modules in contemporary language model architectures possess intrinsic privacy vulnerabilities. 

This paper focuses on Transformer-based language models featuring a distinctive Pooler layer and proposes a two-stage privacy attack method that exploits vulnerabilities in this specific module. Specifically, in the first stage, we employ an analytics-based reconstruction method to initially recover the direction of features associated with this module. This feature information is not averaged over the number of samples in the same batch or the sequence length, enabling the extraction of more unique information for each token. Subsequently, we utilize this feature information as an additional (beyond gradient and priors) supervisory signal to guide the recovery of training data. This research differentiates itself in the following ways: \textbf{\setcounter{myCounter}{1}\roman{myCounter}}) Different from \citet{wang2023reconstructing}, the research is solidly based on textual data and deep language models in real-world scenarios. \textbf{\setcounter{myCounter}{2}\roman{myCounter}}) Different from \citet{fowl2022decepticons} and \citet{boenisch2023curious}, this research does not depend on trap weights, such as Identity module, and maintains adherence to an effective training roadmap. \textbf{\setcounter{myCounter}{3}\roman{myCounter}}) In contrast to the honest-but-curious works~\citep{zhu2019deep, deng2021tag, balunovic2022lamp, gupta2022recovering}, this research provides feature-level supervisory signals that differ from conventional gradients and priors.

The contributions of this research are outlined as follows: 
\textbf{1}) Instead of directly recovering the training samples of the entire model, this paper proposes a two-stage attack method to first approximate the intermediate feature information and then recover the real input. 
\textbf{2}) We design a strategic weight initialization method coupled with a flexible tuning approach, thereby empowering an analytics-based method to accurately and efficiently deduce the direction of intermediate features in a specific module.
\textbf{3}) When integrated with gradient inversion and prior knowledge, our method consistently surpasses previous methods in the attack performance across a variety of benchmark datasets and scenarios. We also propose using the distance between semantic embeddings as a more comprehensive metric for evaluation.
\textbf{4}) This research brings to light the inherent privacy leakage risks that are embedded within the design of contemporary language model architectures. 

\section{Preliminaries}
In this section, we describe the relevant background of federated learning, gradient inversion, priors knowledge, two-layer-networks-based reconstruction, as well as the threat model of our proposed attack method.

\subsection{Federated Learning}
Introduced by~\citet{mcmahan2017communication}, federated learning solves data privacy concerns by promoting decentralized model training. In this approach, models are refined using local updates from individual clients, which are merged at a central server~\citep{konečný2015federated, konečný2016federated, konečný2017federated}. This field has attracted significant attention due to its potential business applications, underlining its promise in academia and industry~\citep{ ramaswamy2019federated, li2020federated}.

\subsection{Gradient Inversion}

Gradient inversion is a significant technique that could potentially breach privacy in federated learning~\citep{ zhu2019deep, zhao2020idlg}. Although federated learning is designed to provide a decentralized training mechanism ensuring local data privacy, gradient inversion shows that this privacy may not be infallible.

\textbf{Problem definition:} 

Consider the supervised learning framework wherein a neural network $f(\cdot; \bm{\Theta}) : \bm{x} \in \mathbb{R}^d \rightarrow f(\bm{x}; \bm{\Theta}) \in \mathbb{R}$ is trained using the objective:
\begin{equation}
    \min _{\bm{\Theta}} \sum_{(\bm{x}, y) \in \mathcal{D}} \ell(f(\bm{x} ; \bm{\Theta}), y)
\end{equation}
where $\ell$ is the loss function and $\mathcal{D}$ denotes the dataset of input-output pairs. In the federated paradigm, during the communication between the central server and clients, each node reports an average gradient of its local batch $S$~\citep{mcmahan2017communication, konečný2015federated}. This is mathematically formulated as:
\begin{equation}
    G := \frac{1}{B} \nabla_{\bm{\Theta}} \sum_{i=1}^B \ell\left(f\left(\bm{x}_i, \bm{\Theta}\right), y_i\right)
\end{equation}
where $B$ is the batch size of $S$. Given the above definition, the challenge posed by gradient inversion becomes apparent: With access to a once-queried gradient and a known model as well as its loss function, is it possible to reconstruct the input training data?

\textbf{Objective of Gradient Inversion:} 
Based on the above problem, the objective of gradient inversion can be represented as:
\begin{equation}
    \min_{\{\hat{\bm{x}}_i, \hat{y}_i\}_{i=1}^B} d\left(\frac{1}{B} \sum_{i=1}^B \nabla_{\bm{\Theta}} \ell\left(f\left(\hat{\bm{x}}_i ; \bm{\Theta}\right), \hat{y}_i\right), G\right)
\end{equation}
Here, $d(\cdot, \cdot)$ quantifies the difference between the provided and deduced gradients, and ($\hat{x_I}, \hat{y_i}$) refers to the estimated input and its label. Prominent works have leveraged this objective to attempt the retrieval of private data~\citep{zhu2019deep, zhao2020idlg, geiping2020inverting}.

\subsection{Prior Knowledge}

Relying solely on gradient inversion to recover textual data often proves challenging, especially when handling larger batch sizes and long sequences. Researchers often seek to acquire the prior knowledge encapsulated in pre-trained language models like GPT-2~\citep{radford2019language} to address this. These models are adept at predicting the probability of the next token based on the preceding sequence. This property aids in evaluating the quality of text searched through gradient inversion. Specifically, it introduces the perplexity as an evaluation metric to guide optimization by pinpointing optimal starting points or intermediate points of the attack~\citep{balunovic2022lamp, gupta2022recovering}.

\subsection{Two-layer Network-based Reconstruction}\label{section:3-3}

\citet{wang2023reconstructing} identified a gap in existing literature regarding the capability of gradient information to unveil training data. Their study 
demonstrates that it might be possible to reconstruct training data solely from gradient data using a theoretical approach within a two-layer neural network. Consider a two-layer neural network:
$f(x;\Theta) = \sum_{j=1}^{m} a_j \sigma(w_j \cdot x)$,
with parameters defined as \( \Theta = (a_1, ... , a_m, w_1, ... , w_m) \). Here, $m$ represents the hidden dimension. The objective function is represented as:
$L(\Theta) = \sum_{i=1}^{B} (y_i - f(x_i;\Theta))^2$.
A notable finding is that the gradient for $a_j$ is solely influenced by $w_j$, making it independent from other parameters. This gradient is represented as:
\begin{equation}
g_{j}:=\nabla_{a_{j}}L(\Theta)=\sum_{i=1}^{B}r_{i}\sigma\left(w_{j}^{\mathsf{T}}x_{i}\right)
\end{equation}
where the residual \(r_{i}\) is given by \(r_{i}=f(x_{i};\Theta)-y_{i}\). For wide neural networks with random initialization from a standard normal distribution, the residuals \( r_{i} \) concentrate to a constant, \( r^{*}_{i} \). By set $g_{(w)}:=\sum_{i=1}^{B}r_{i}^{*}\sigma(w^{\top}x_{i})$, \(g_{j}\) can be expressed as \(g_{j} = g(w_{j}) + \epsilon\), where \(\epsilon\) represents noise. Then the third derivative of \(g(w)\) is represented as:
\begin{equation}
\nabla^{3}g(w) = \sum_{i=1}^{B} r_{i}^{*} \sigma^{(3)}(w^{\textsf{T}} x_{i}) x_{i}^{\otimes3}
\label{equation/collapse}
\end{equation}
Here, $x_{i}^{\otimes3}$ signifies the tensor product of vector $x_{i}$ with itself three times. The researchers postulated that if they can accurately estimate \(\nabla^{3}g(w)\), it is possible to determine \(\{x_{i}\}_{i=1}^{B}\) by using tensor decomposition techniques, especially when these features are independent. They used Stein’s Lemma, expressed as:
$\mathbb{E}[g(X)H_{p}(X)] = \mathbb{E}[g^{(p)}(X)]$
to approximate \(\nabla^{3}g(w)\) as:
\begin{align}
T &= \mathbb{E}_{W}[\nabla_{W}^{3}g(W)] = \mathbb{E}_{W\sim N(0,I)}[g(W)H_{3}(W)] \nonumber \\
  &\approx \frac{1}{m} \sum_{j=1}^{m} g(w_{j})H_{3}(w_{j}) = \hat{T} \label{equation/concentration}
\end{align}
Where $H_{3}(w_j)$ is the p-th Hermite function of $w_j$. By leveraging this approach, they successfully reconstructed each unique $x_{i}$. Their approach is primarily theoretical, focusing on two-layer fully connected networks and largely confined to randomly initialized networks. When applied to deeper networks, their method uses identity modules and other transparently detectable weight manipulations, which also limits its practical use. 

Inspired by \citet{wang2023reconstructing}, we have identified a vulnerable module commonly found in current language model architectures, such as BERT and RoBERTa~\citep{devlin2018bert, liu2019roberta}. The module, comprising the Pooler and Classifier layers, typically forms the final stages of these networks. Given this module does not have deeper topological connections, it can be treated as an independent two-layer network. Leveraging this insight, we have designed specific techniques to enhance the two-layer network-based reconstruction method, enabling it to recover the feature information associated with this special module in pre-trained language models. Further details of this methodology are discussed in Section~\ref{section:4-2}.

\subsection{Threat Model\label{section:3-5}}

To facilitate comparisons with previous works, our threat model is based on the following principles: \textbf{1)} We opt to freeze the gradients of both token and positional embeddings. This is because it is relatively easy to deduce the training text tokens or max sequence length without this restriction since only the tokens and positions in the current training batch will receive gradient updates on the embedding matrix. By freezing the updates of these embeddings, we create a more challenging attack scenario. \textbf{2)} We ensure that the training process maintains effective gradient aggregation and consistently aims to minimize the training loss. This principle is crucial for preserving the integrity and efficiency of the training process, ensuring it remains free from suspicion. \textbf{3)} We refrain from using trap weights, such as the Identity module, in our approach. However, instead of using pre-trained weights for initialization, we may opt to initialize randomly for partial weights of certain layers.

\section{Methodology}

Gradient inversion seeks to reconstruct the original training data by harnessing the gradients of a known network. A closer look at this method reveals several challenges. Central to these is the nonconvexity of the issue, marked by the presence of numerous local minima that complicate the pursuit of the global optimum. Additionally, the problem is over-determined because it has more equations to resolve than unknown parameters. While these equations remain consistent, they complicate the optimization process. This complexity persists even when reduced to a single-sample scenario. As a result, gradient inversion remains an NP-complete problem, implying that procuring an exact solution within a feasible time frame is difficult~\citep{wang2023reconstructing}. From a broader perspective, it is crucial to recognize that text tokens represent discrete data, different from the continuous nature of images. Additionally, the gradients of language models are not just averaged over the number of samples in the same batch, but also across the sequence length of the longest sentence in that batch. These properties make the recovery of textual training data more challenging compared to image data. 

To overcome these challenges, this paper endeavors to seek more valuable information that could guide the recovery or retrieval process. Instead of solely focusing on gradients and external information (priors), we shift our attention toward the inner workings of the model itself.

\subsection{Vulnerable Module Identification}

Building upon the work of \citet{wang2023reconstructing}, which demonstrated the feasibility of recovering training data using gradient information from a two-layer fully connected network, we are reminded to scrutinize similar vulnerabilities embedded in contemporary language model architectures. However, the increasing complexity and depth of these models render it nearly impossible to apply a similar strategy without resorting to trap weights, such as an identity module. This leads us to a critical observation when examining the architecture of widely-used transformer-based language models like BERT and RoBERTa. Notably, these models contain a module consisting of a Pooler layer and a Classifier layer, with a non-linear activation function between them. By analyzing the topological operational order, we discern that this module can be treated as an independent two-layer fully connected network, given its position at the top of the language model, implying no deeper logical operations. 

With this keen realization, instead of attempting to recover the training samples of the entire model in one step, we propose first recovering the input information for this special module using an analytics-based method. This additional information then aids the optimization-based method (gradient inversion) with extra supervisory signals, thereby enhancing privacy attack performance. we have successfully unearthed a novel source of valuable information, distinct from conventional gradients and priors, embedded within the internal module of the model itself. We identify this module as a \textbf{vulnerable module}, which harbors inherent privacy leakage risks. Nonetheless,~\citet{wang2023reconstructing}'s reconstruction method, as described in Section ~\ref{section:3-3}, comes with its own set of assumptions and limitations. Most critically, it does not recover the actual features but their direction within the feature space. To ensure the efficacy of this analytics-based method, especially in its application to real pre-trained language models, we have designed several critical techniques to relax or remove these limitations.

\subsection{First-stage Analytics-based Attack \label{section:4-2}}

In the first-stage attack, we aim to first recover the feature information directed to the identified vulnerable module with an enhanced analytics-based method. To better elucidate our tailored design, let us first establish the notations used in this context: Let $X \in \mathbb{R}^{B \times d}$ be the input to this vulnerable module, where $B$ is the batch size and $d$ is the feature dimension. Let $W_1 \in \mathbb{R}^{d \times v}$ and $W_2 \in \mathbb{R}^{v \times n}$  denote the weights of the Pooler layer and the Classifier layer, respectively, with $v$ being the pooler dimension and $n$ being number of classes. Let $\sigma$ represent the non-linear activation function that is positioned after the Pooler layer. 

\textbf{Enlarge Pooler Dimension:~\label{section:4-2-1}} 
The initial configuration of language models often sets the pooler dimension $v$ to match the hidden dimension $d$ (For BERT$_{\text{BASE}}$, it's 768)~\citep{devlin2018bert}. This setting is insufficient to promise the accuracy of tensor decomposition when applying the analytics-based reconstruction method. To address this limitation, we temporarily expand the pooler dimension to match the vocabulary size for BERT. To be explicit, we set $v = |V|$. The rationale behind this change is grounded in enhancing the model's expressiveness while ensuring our modification is subtle. It's important to note that, given the expanding width of state-of-the-art language models such as GPT-3~\cite{brown2020language}, the hidden dimension has become sufficiently large to assure the accuracy of our method. Consequently, we can bypass this certain step that would otherwise be necessary in BERT.

\textbf{Strategic Weight Initialization:~\label{section:4-2-2}}
As mentioned in Section~\ref{section:3-3}, $m$ signifies the intermediate dimension in the two-layer network. In our setting, we should have $m = v$. However, during our computation of $\hat{T}$ as outlined in Equation~\ref{equation/concentration}, we noticed an anomaly in $g_j$. Due to the random initialization of $W_1$ by requirement, a substantial portion of gradients $g_j$ approached a value close to 0. This side effect impacts the subsequent decomposition procedure. To address this issue, rather than setting $m = v$, we set $m = v - d$. This approach ensures the original pre-trained weights $W_1^\prime \in \mathbb{R}^{d \times d}$ are retained in the new weight matrix $W_1 \in \mathbb{R}^{d \times v}$, allowing us to obtain optimal gradients for $W_1$ and $W_2$. Simultaneously, the remaining dimensions $W_1^{\prime\prime} \in \mathbb{R}^{d \times (v-d)}$ are randomly initialized and adequate to promise the accuracy of tensor decomposition. For the classifier layer, we utilize a strategy similar to that of the Pooler layer, adjusting the remaining dimensions to a constant ($i/m$, where $i$ represents the class index for the classification task). More details are in Appendix~\ref{B-1}.

From one perspective, it may appear that we have altered the parameters. However, it is important to clarify that we have not assigned any special properties to these weights. Our approach involves initializing part of the weight randomly, which is a standard operation in model initialization. Furthermore, this random initialization is confined only to the identified vulnerable module, allowing the rest of the language model to utilize pre-trained weights for initialization. Consequently, this approach avoids creating trap weights and ensures the normal training roadmap.

\textbf{Flexible Tuning Framework:~\label{section:4-2-3}}
\citet{wang2023reconstructing} suggests significantly expanding the pooler dimension $m$ in comparison to the input dimension $d$ to reduce the tensor decomposition error. In our setting, the specific relationship between recovery dimension $d$ and pooler dimension $v = m + d$ remains undetermined. Recognizing these constraints, we keep $m$ constant and design an alternative method to tweak \(d\). Specifically, instead of attempting to recover the full dimension $d$, our strategy focuses on recovering a dimension $d^{\prime}$ and $d^{\prime} \leq d$. This approach sets the subweights~($d\colon, d^{\prime}\colon$) of $W_1$ to zero. Then the gradient $g_j$ in Equation~\ref{equation/concentration} remains functional but is exclusively tied to the subweights~($\colon,\colon{d^\prime}$) of $W_1$. As a result, we embrace a more flexible and efficient methodology by centering our reconstruction on the feature subset~$X \in \mathbb{R}^{B \times d^\prime}$. More details can be found in Appendix~\ref{B-1}.

\textbf{Reorder Feature Information:~\label{section:4-2-4}}
When applying tensor decomposition techniques to retrieve features from $\hat{T}$, a significant issue arises when the batch size exceeds one: the exact order of the recovered features remains uncertain. Under adversarial conditions, one might try every conceivable permutation as a reference. However, we simplify the procedure by sequentially comparing each recovered feature to the actual input features with cosine similarity until the best order is discerned. In certain cases, a single recovered feature displayed a notably high cosine similarity with multiple actual inputs simultaneously. Interestingly, although a 1-m greedy relationship might exhibit a high correlation, it did not exceed the attack performance of a straightforward 1-1 match in the final outcome. Consequently, we adopted the 1-1 relationship to achieve the best attack result.

\textbf{Activation Function Exploration:~\label{section:4-2-5}}
Our empirical observations have indicated that various activation functions result in differing effects on information retrieval. Consequently, rather than limiting our focus to the original activation function, we have experimented with a range of activation functions to assess their impact on the final attack performance. Currently, we include activation functions such as Tanh, ReLU, SeLU, and a custom function defined as $\sigma(x) = x^3 + x^2$. A more detailed discussion of this design will be presented in Section~\ref{section:5-3}.

\subsection{Second-stage Optimization-based Attack~\label{section:4-3}}

In the second-stage attack, we aim to recover the real input for the entire language models with an optimization-based method. Specifically, following \citet{balunovic2022lamp}, we divide our entire optimization process into three phases: Initialization, Training, and Token Swap. In the initialization and token swap stages, we aim to leverage certain metrics to identify optimal starting or intermediary points for the subsequent training phase. This stage is also commonly recognized as discrete optimization. 
In this setting, we've chosen a mix of metrics to guide the choice, including gradient match loss and perplexity obtained from pre-trained language models. 
In the optimization stage, we optimize the embeddings derived from input IDs to minimize gradient matching loss and cosine distance between the input of the Pooler layer with the recovered feature information from our first-stage attack. This phase falls under the category of continuous optimization. We oscillate between continuous and discrete optimization phases to bolster the final attack performance. More details can be found in Appendix~\ref{B-1}.

\section{Experiments}

This section initially presents the fundamental setup for our experiments. Subsequently, we demonstrate the results of experiments in various settings and provide an in-depth analysis from multiple perspectives.

\subsection{Set Up~\label{section:5-1}}
\textbf{Datasets:}
Following previous work~\cite{balunovic2022lamp}, our experimental design incorporates three binary text classification datasets to ensure a comprehensive evaluation. Specifically, we utilize \textbf{CoLA} and \textbf{SST-2} from the \textbf{GLUE} benchmark~\citep{warstadt2018neural, socher2013recursive, wang2019glue}, with their sequences predominantly ranging between 5-9 and 3-13 words, respectively. Additionally, the~\textbf{RottenTomatoes} dataset presents a more complex scenario with sequence lengths oscillating between 14 and 27 words~\citep{pang2005seeing}. More details can be found in Appendix~\ref{sec:C-3}. Within the scope of our experiments, we utilize a subset of 100 randomly selected sequences from the training sets of these datasets as our evaluation benchmark, a method also endorsed by~\citet{balunovic2022lamp}.

\textbf{Models:} 
Experiments are primarily based on \textbf{BERT$_{\text{BASE}}$} \citep{devlin2018bert} architecture. Consistent with~\citet{balunovic2022lamp}, we use models that have been fine-tuned on downstream tasks for two epochs. To ensure a fair comparison, we adopt the same fine-tuned models from~\citet{balunovic2022lamp}. As for the auxiliary language model employed to extract prior knowledge, we choose GPT-2~\citep{radford2019language}, a choice also used by~\citet{balunovic2022lamp}.

\textbf{Metrics:}
Following~\citet{deng2021tag} and~\citet{balunovic2022lamp}, we evaluate attack performance with the ROUGE metric suite~\citep{lin2004rouge}. Specifically, we present the F-scores of \textbf{ROUGE-1}, \textbf{ROUGE-2}, and \textbf{ROUGE-L}. These metrics respectively assess the retrieval of unigrams, bigrams, and the proportion of the longest continuous matching subsequence relative to the entire sequence. We omit all padding tokens in the reconstruction and evaluation phases.

\textbf{Baselines:}
We benchmark our approach against three baselines: \textbf{DLG}, \textbf{TAG}, and \textbf{LAMP}, with a similar definition of threat model described in Section~\ref{section:3-5}. Among them, LAMP represents the state-of-the-art. We employ the open-sourced implementation from LAMP, which encompasses the implementations for all three baselines~\citep{deng2021tag, zhu2019deep, balunovic2022lamp}. Following previous work, we assume the lengths of sequences are known for both baselines and our attacks, as an adversary can run the attack for all possible lengths~\citep{balunovic2022lamp}.

\textbf{Implementation:}
Our method is implemented based on LAMP's framework. To ensure a fair comparison, we standardized the experimental conditions and settings when comparing our approach with baselines. We adopt all of LAMP's hyperparameters, including the optimizer, learning rate, learning rate schedule, regularization coefficient, and optimization steps. 
For hyperparameters unique to our method, we made selections using a grid search on BERT$_{\text{BASE}}$ and shared them in different settings. We also assume the prior knowledge of input labels. This is typical in text classification tasks with limited categories because we can easily iterate all the potentials. More details can be found in Appendix~\ref{B-1}.

\renewcommand{\arraystretch}{1.2}
\begin{table*}[ht]
\tiny
\centering
\caption{Analysis of Text Privacy Attacks on BERT$_{\text{BASE}}$: Impact of Various Batch Sizes and Datasets. The terms R-1, R-2, and R-L represent ROUGE-1, ROUGE-2, and ROUGE-L scores, respectively. We calculate the average change by comparing it with the baseline LAMP$_{\text{COS}}$. Symbols $\downarrow$ and $\uparrow$ denote degradation and improvement, respectively. Notably, a bold format is used to highlight the most effective attack performance under identical settings. The symbol $\star$ signifies the model's original activation function.}
\begin{tabularx}{\textwidth}{l|CCC|CCC|CCC|CCC|CCC}
\hline
Method & \multicolumn{3}{c|}{B=1} & \multicolumn{3}{c|}{B=2} & \multicolumn{3}{c|}{B=4} & \multicolumn{3}{c|}{B=8} & \multicolumn{3}{c}{Avg~$\Delta$} \\
 & R-1 & R-2 & R-L & R-1 & R-2 & R-L & R-1 & R-2 & R-L & R-1 & R-2 & R-L & R-1 & R-2 & R-L \\
\hline
\hline
\multicolumn{13}{l}{\color{blue}CoLA} \\
\hline
DLG & 59.3 & 7.7 & 46.2 & 36.9 & 2.6 & 31.4 & 35.3 & 1.4 & 31.9 & 16.5 & 0.8 & 7.9 & $\downarrow$~17.7 & $\downarrow$~16.8 & $\downarrow$~19.0 \\
TAG & 78.9 & 10.2 & 53.3 & 45.6 & 4.6 & 36.9 & 35.3 & 1.6 & 31.3 & 33.3 & 1.6 & 30.4 & $\downarrow$~6.4 & $\downarrow$~15.4 & $\downarrow$~10.4 \\
LAMP$~_{\text{COS}}$ & 84.8 & 46.2 & 73.1 & 57.2 & 21.9 & 49.8 & 40.4 & 6.4 & 36.2 & 36.4 & 5.1 & 34.4 & --- & --- & --- \\
\hline
\textbf{Ours}$_{\text{~Tanh}^{\star}}$ & 84.5 & 46.1 & 72.8 & 56.9 & 22.0 & 49.6 & 41.2 & 7.8 & 40.1 & 37.2 & 5.2 & 34.4 & $\uparrow$~0.25 & $\uparrow$~0.4 & $\uparrow$~0.9 \\
\textbf{Ours}$_{\text{~ReLU}}$ & 84.5 & 45.9 & 72.6 & 57.3 & 19.3 & 49.8 & 42.3 & 8.4 & 40.1 & 37.6 & 5.6 & 34.5 & $\uparrow$~0.7 & $\downarrow$~0.1 & $\uparrow$~0.9 \\
\textbf{Ours}$_{\text{~SeLU}}$ & \textbf{86.6} & \textbf{51.5} & \textbf{76.7} & \textbf{69.5} & \textbf{31.2} & \textbf{60.6} & \textbf{50.5} & \textbf{11.8} & \textbf{43.9} & \textbf{40.8} & \textbf{8.3} & \textbf{38.1} & \textbf{$\uparrow$~7.1} & \textbf{$\uparrow$~5.8} & \textbf{$\uparrow$~6.5} \\
\textbf{Ours$_{~x^3+x^2}$} & 84.6 & 45.2 & 72.4 & 57.3 & 19.2 & 49.8 & 43.9 & 11.4 & 40.1 & 37.8 & 5.9 & 34.8 & $\uparrow$~1.2 & $\uparrow$~0.5 & $\uparrow$~1.0 \\
\hline
\multicolumn{13}{l}{\color{blue}SST-2} \\
\hline
DLG & 57.7 & 11.7 & 48.2 & 39.1 & 7.6 & 37.2 & 38.7 & 6.5 & 36.4 & 36.6 & 4.7 & 35.5 & $\downarrow$~16.0 & $\downarrow$~19.3 & $\downarrow$~14.1 \\
TAG & 71.8 & 16.1 & 54.4 & 46.1 & 10.9 & 41.6 & 44.5 & 9.1 & 40.1 & 41.4 & 6.7 & 38.9 & $\downarrow$8.0 & $\downarrow$16.2 & $\downarrow$9.7 \\
LAMP$~_{\text{COS}}$ & 87.7 & 54.1 & 76.4 & 59.6 & 26.5 & 53.8 & 48.9 & 17.1 & 45.4 & 39.7 & 10.0 & 38.2 & --- & --- & --- \\
\hline
\textbf{Ours}$_{\text{~Tanh}^{\star}}$ & 88.5 & 56.9 & 77.3 & 66.4 & 33.2 & 61.2 & 49.9 & 15.6 & 46.1 & 43.5 & 10.9 & 40.5 &  $\uparrow$~3.1 & $\uparrow$~2.2 & $\uparrow$~2.8 \\
\textbf{Ours}$_{\text{~ReLU}}$ & 88.5 & 56.1 & 77.1 & 67.3 & 32.6 & 60.8 & 50.4 & 14.1 & 46.2 & 43.5 & 11.2 & 40.8 & $\uparrow$~3.4 & $\uparrow$~1.6 & $\uparrow$~2.8 \\
\textbf{Ours}$_{\text{~SeLU}}$ & 90.3 & 59.0 & 78.2 & 71.0 & 35.3 & 63.4 & 58.6 & \textbf{26.3} & 54.2 & 45.4 & 11.5 & 43.2 & $\uparrow$~7.3 & $\uparrow$~6.1 & $\uparrow$~6.3 \\
\textbf{Ours$_{~x^3+x^2}$} & \textbf{93.1} & \textbf{61.6} & \textbf{81.5} & \textbf{78.3} & \textbf{40.9} & \textbf{67.9} & \textbf{60.6} & 23.1 & \textbf{54.9} & \textbf{49.5} & \textbf{16.5} & \textbf{47.3} & \textbf{$\uparrow$~11.4} & \textbf{$\uparrow$~8.6} & \textbf{$\uparrow$~9.5} \\
\hline
\multicolumn{13}{l}{\color{blue}Rotten Tomatoes} \\
\hline
DLG & 20.1 & 0.4 & 15.2 & 18.9 & 0.6 & 15.4 & 18.7 & 0.4 & 15.7 & 20.0 & 0.3 & 16.9 & $\downarrow$~17.4 & $\downarrow$~5.4 & $\downarrow$~11.5 \\
TAG & 31.7 & 2.5 & 20.1 & 26.9 & 1.0 & 19.1 & 27.9 & 0.9 & 20.2 & 22.6 & 0.8 & 18.5 & $\downarrow$~9.5 & $\downarrow$~4.5 & $\downarrow$~7.8 \\
LAMP$~_{\text{COS}}$ & 63.4 & 13.8 & 42.6 & 38.4 & 6.4 & 28.8 & 24.6 & 2.3 & 20.0 & 20.7 & 0.7 & 17.7 & --- & --- & --- \\
\hline
\textbf{Ours}$_{\text{~Tanh}^{\star}}$ & 64.2 & 15.5 & 43.8 & 38.8 & 5.8 & 28.9 & 28.3 & 2.4 & 21.2 & 22.4 & 1.1 & 18.9 & $\uparrow$~1.7 & $\uparrow$~0.4 & $\uparrow$~1.0 \\
\textbf{Ours}$_{\text{~ReLU}}$ & 64.1 & 15.7 & 44.2 & 40.2 & 5.4 & 28.8 & 31.1 & 2.6 & 23.6 & 22.8 & 1.3 & 18.8 & $\uparrow$~2.8 & $\uparrow$~0.5 & $\uparrow$~1.6 \\
\textbf{Ours}$_{\text{~SeLU}}$ & 71.9 & 19.2 & 48.7 & \textbf{48.1} & \textbf{8.2} & \textbf{34.2} & \textbf{33.0} & \textbf{4.23} & \textbf{25.3} & \textbf{24.6} & \textbf{2.0} & \textbf{20.6} & \textbf{$\uparrow$~7.6} & \textbf{$\uparrow$~2.6} & \textbf{$\uparrow$~4.9} \\
\textbf{Ours$_{~x^3+x^2}$} & \textbf{72.2} & \textbf{21.0} & \textbf{49.3} & 44.6 & 7.0 & 31.8 & 29.9 & 3.5 & 24.3 & 23.6 & 1.7 & 19.8 & $\uparrow$~5.8 & $\uparrow$~2.5 & $\uparrow$~4.0\\
\hline
\hline
\end{tabularx}
\label{tab:1}
\end{table*}

\subsection{Results and Analysis}

We present experimental results in Table~\ref{tab:1}. These findings demonstrate that our approach outperforms all baselines~(DLG, TAG, and LAMP) across various datasets and batch sizes. Examining the impact of batch size variations, we notice that launching an attack becomes more challenging as the batch size increases. All attack methods, including ours, exhibit a decline in attack performance. However, our method brings a more noticeable improvement at batch sizes 2 and 4, surpassing its efficacy at batch sizes 1 and 8. We posit that for a batch size of 1, where the gradient is only averaged solely over tokens, the benefit of incorporating the feature information is less evident because the gradient information still plays a leading role in the optimization process. For a batch size of 8, the improvement scale is also not pronounced, we explore the background reason in Section~\ref{section:5-3}.

\begin{wrapfigure}{r}{0.5\textwidth}
  \centering
  \includegraphics[width=\linewidth]{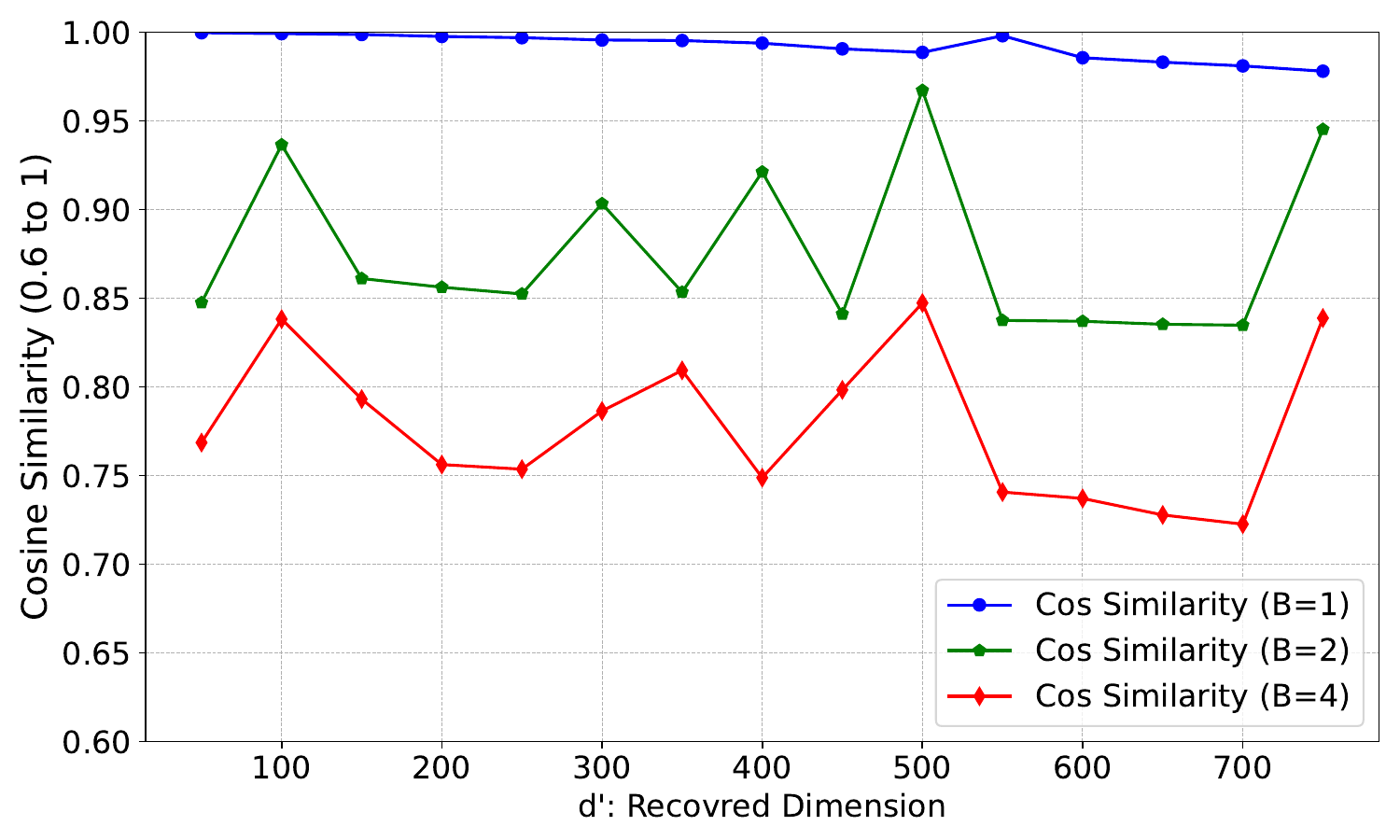}
    \captionof{figure}{Cosine similarity between recovered features and ground truth of BERT$_{\text{BASE}}$ on SST-2 across varying dimensions (50$\sim$750 in 50-step intervals) and batch sizes (1, 2, 4)}
    \label{fig:2}
\end{wrapfigure}

Turning our attention to variations in sequence length across datasets, we notice a clear trend: as sequences get longer, the benefit from feature information at a batch size of 1 becomes more pronounced. Specifically, for the CoLA dataset with token counts between 5-9, we see an average improvement in ROUGE metrics of \textbf{3\%}. This improvement grows to \textbf{5\%} for the SST-2 dataset with token counts from 2 to 13. For the Rotten Tomatoes dataset, which features even longer sequences with token counts ranging from 14 to 27, the average ROUGE metric improvement further increases to \textbf{8\%}. This suggests a correlation between sequence length and the extent of improvement observed. Moreover, when the batch size exceeds one, the benefits observed for these three datasets are consistently notable. Recall that gradient averaging occurs only over tokens at a batch size of 1, it implies that with longer sentences, the gradient information becomes less effective, leading to greater benefits from feature-level supervision signals. When batch sizes are larger than 1, averaging happens over the number of tokens and sentences simultaneously. This results in our method consistently yielding pronounced benefits across sequences with different lengths. Our findings further reinforce the idea that relying exclusively on gradient information diminishes efficacy with larger batch sizes and longer sequences.

In a word, the experiments consistently show that the use of feature information significantly enhances the success of privacy attacks on language models. This evidence further confirms that current language models, particularly in their Pooler and Classifier layers, inherently possess vulnerabilities that pose risks of privacy leakage.

\section{Discussion}\label{section:5-3}

\textbf{Impact of Activation Function:} As outlined in Section~\ref{section:4-2-5}, we replaced the original Tanh activation function with ReLU, SeLU, and a custom activation function $\sigma=x^3+x^2$ to investigate how different activation functions affect attack performance in our strategy. Table~\ref{tab:1} presents the performance of these attacks under various settings. SeLU and $\sigma=x^3+x^2$ consistently yield significant improvements in attack performance, while the enhancements seen with Tanh and ReLU are relatively less pronounced. We speculate that the latter activation functions' n-th derivative leads to a zero expectation ($\mathbb{E}_{Z\sim N(0,1)}[\sigma^{(n)}(Z)]=0$), thereby affecting the estimation of $T$ as explained in Equation~\ref{equation/concentration}. In contrast, the earlier activation functions, whose n-th derivatives are neither odd nor even, do not exhibit this issue, potentially resulting in a more pronounced risk of privacy leakage. Another interesting phenomenon observed is that for the CoLA dataset, all attack methods show only minor improvements, except for attacks utilizing the SeLU. This suggests the existence of an unknown correlation between datasets and the feature information recovered in different activation functions. Further details and discussions are provided in the Appendix~\ref{B-3}.

\textbf{Impact of Recovery Dimension:} In Section~\ref{section:4-2-3}, we propose fixing $m$ and adjusting $d'$ to identify the optimal mapping for $d'$ (where $d' < d$) and $m$. Accordingly, we conduct experiments using \textbf{BERT$_{\text{BASE}}$} with various batch sizes to investigate the quality of the recovered feature information by calculating their cosine similarity with the ground truth. The results are illustrated in Figure~\ref{fig:2}. Our findings suggest that when the batch size is 1, the recovered quality gradually degrades as the recovery dimension $d'$ increases, yet it remains as high as 0.99 across all configurations. However, this pattern does not hold when the batch size exceeds 1. We also observed that the recovered quality consistently declines as the batch size increases. We hypothesize that multiple inputs might exhibit some undisclosed dependencies, particularly features within the deeper layers of language models, thereby affecting the efficacy of tensor decomposition. For simplicity, we set $d'=100$ across all experiments. However, under adversarial conditions, attackers might experiment with various $d'$ settings to enhance their attack performance.

\textbf{Impact on Other Models:} To demonstrate the effectiveness of our attack method on various model architectures, we also apply our method on the RoBERTa~\citep{liu2019roberta}. While RoBERTa shares similarities with BERT, it distinguishes itself through unique training configurations and datasets. Notably, unlike BERT$_{\text{BASE}}$, RoBERTa does not have a Pooler layer. Instead, it employs a classifier composed of two linear layers in the head. In our experiments, we treat the first layer as an analogous Pooler layer and endeavor to reconstruct its input first. All the models used in this experiment are from Hugging Face, contributed by TextAttack~\citep{morris2020textattack}. As for the auxiliary model, we employ RoBERTa itself due to a specific challenge: we can't locate another generative model using the same tokenizer with RoBERTa. However, it's essential to note that we use the exact same settings for baselines and our method. We present the experiment results in Table~\ref{tab:3}. While the overall attack performance significantly decreases due to the auxiliary masked language model, our approach still outperforms the baseline.
Furthermore, in numerous instances (as illustrated in Table~\ref{tab:3}), our method appears to restore the essence of the reference sample almost flawlessly. However, due to the limitation of evaluation metrics, they may have equal or even worse evaluation metrics than some terrible cases. Hence, we also employ the cosine similarity metric between the embeddings generated by SBERT for both the reference and recovery texts to assess the attack performance~\citep{reimers2019sentence}. 

\renewcommand{\arraystretch}{1.2}
\begin{table*}[ht]
\small
\centering
\caption{Text privacy attack on RoBERTa$~_{\text{BASE}}$. R-1, R-2, and R-L are same within Table~\ref{tab:1}. Cos$_{\text{S}}$ indicates the average cosine similarity between references and recovered samples.}
\begin{tabularx}{\textwidth}{l|p{1cm}|CCCC|p{6.5cm}}
\hline
Dataset & Method & R-1 & R-2 & R-L & Cos$_{\text{S}}$ & Recovered Samples \\
\hline
\hline
\multirow{3}{*}{CoLA} & \multicolumn{6}{c}{reference sample: \textcolor{blue}{The box contains the ball}} \\ \cline{2-7}
& LAMP & 15.5 & 2.6 & 14.4 & 0.36 & likeTHETw box contains divPORa \\
& \textbf{Ours} & \textbf{17.4} & \textbf{3.8} & \textbf{15.9} & \textbf{0.41} & \textcolor{orange}{like Mess box contains contains balls} \\
\hline
\multirow{3}{*}{SST2} & \multicolumn{6}{c}{reference sample: \textcolor{blue}{slightly disappointed}} \\ \cline{2-7}
& LAMP & \textbf{20.1} & \textbf{2.2} & 15.9 & 0.56 & likesmlightly disappointed a \\
& \textbf{Ours} & 19.7 & 2.1 & \textbf{16.8} & \textbf{0.59} & \textcolor{orange}{like lightly disappointed a} \\
\hline
\multirow{3}{*}{Toma} & \multicolumn{6}{c}{reference sample: \textcolor{blue}{vaguely interesting, but it's just too too much}} \\ \cline{2-7} 
& LAMP & 19.9 & 1.6 & 15.1 & 0.48 & vagueLY', interestingtooMuchbuttoojusta \\
& \textbf{Ours} & \textbf{21.5} & \textbf{1.8} & \textbf{16.0} & \textbf{0.51} & vagueLY, interestingBut seemsMuch Toolaughs \\
\hline
\end{tabularx}
\label{tab:3}
\end{table*}

\section{Related Work\label{section-2-2}}

While federated learning features with data privacy, recent studies show that model updates (gradients and parameters) can be intentionally leveraged to uncover sensitive data~\citep{phong2017privacy, zhao2020idlg, zhu2020r, zhu2019deep}. This susceptibility is especially pronounced in the field of CV~\citep{huang2021evaluating, geiping2020inverting, yin2021see, jeon2021gradient}.

Textual data poses unique challenges in the context of private data attacks, especially given the prevalence of Transformer architectures. In Transformer, gradients average across sequences and tokens, which inherently masks specific token details. Furthermore, the inputs, expressed as discrete token IDs, starkly contrast the continuous features found in image data. Nonetheless, numerous studies have highlighted the risks associated with textual information. 
\citet{fowl2021robbing, fowl2022decepticons, boenisch2023curious} distribute networks with embedded backdoors or trap parameters that facilitate easy reconstruction of training data. However, one can employ prefixed, recognized architectures to counter the former attack and guard against potential backdoor threats. For the latter attack, consistently monitoring statistics of features and weights across different layers can help detect malicious parameter~\citep{balunovic2022lamp}.
Some approaches assume a trustworthy central server. Even with its integrity, the shared parameters and gradients could still be leveraged to extract private data~\citep{zhu2019deep}. For example, methods introduced by~\citet{zhu2019deep} and~\citet{deng2021tag} employ optimization-based strategies using finely-tuned objective functions for data retrieval.~\citet{balunovic2022lamp} leverages prior knowledge from extensive language models for data recovery. However, due to the self-imported limitation (Server is benign without doing any change to the model), these methods tend to be less effective with larger batch sizes. Notably, the method introduced by~\citep{gupta2022recovering} remains effective even with considerable batch sizes. Nevertheless, this vulnerability can be easily defended by suspending updates of embedding matrix.

\section{Conclusion}

This paper introduces a two-stage privacy attack methodology that exposes the inherent privacy leakage risks in contemporary language model architectures. Rather than attempting to directly reconstruct the training samples of the language model, our approach focuses initially on retrieving certain feature information. Furthermore, our method distinguishes itself by not solely depending on gradients and prior knowledge; it also integrates unique feature-level data. Extensive empirical research, spanning a range of model architectures, datasets, and batch sizes, corroborates the efficacy of our approach.

\section*{Acknowledgments}
The authors would like to extend their sincere gratitude to the ARC (A Root Cluster for Research into Scalable Computer Systems) at the Computer Science Department of North Carolina State University. The invaluable computing resources provided by the ARC cluster (\href{https://arcb.csc.ncsu.edu/~mueller/cluster/arc/}{https://arcb.csc.ncsu.edu/~mueller/cluster/arc/}) were instrumental in facilitating the research presented in this paper.

\newpage

\bibliographystyle{unsrtnat}
\bibliography{arxiv}

\appendix
\newpage
\section{Appendix-A}

As discussed in the main paper, our method distinguishes itself by uncovering the inherent privacy risks embedded in contemporary language model architectures. Additionally, this work also highlights its superior performance in privacy attacks compared to prior research. To more effectively demonstrate this superiority under comparatively fair conditions, we first discuss the difference between previous work and our research, and then clarify our threat model again.

\subsection{Comparision with Previous Work~\label{A-1}}

Previous work classifies privacy attacks on textual data into two principal categories: \textbf{malicious} attacks and \textbf{eavesdropping} attacks, with the latter also known as \textbf{honest-but-curious } attacks~\citep{gupta2022recovering}. However, this binary classification tends to oversimplify the complexity of the issue, failing to account for various nuanced intermediary stages.

Conventionally, \textbf{malicious} attacks are characterized by the assumption of a completely malicious server~\citep{fowl2021robbing,fowl2022decepticons, boenisch2023curious}. Specifically, they may insert specially designed modules into the original model or create trap weights (such as Identity weight) to facilitate the recovery of training samples. Differently, instead of inserting a suspicious module, we opt to identify vulnerable modules that inherently possess privacy risks within contemporary language model architectures. Additionally, our approach does not depend on trap weights; instead, we merely initialize a subset of weights randomly in a specific layer.

\textbf{Malicious} attacks fundamentally forget the primary objective of federated learning, which is to minimize training loss and develop an effective model. In contrast, our method does not impede the original training roadmap due to its effective gradient generation and weight updates.

\textbf{Honest-but-curious} attacks can extract training samples by utilizing only gradient information and prior knowledge. However, these attacks typically demonstrate limited effectiveness, particularly in scenarios using a large batch size. To address this limitation, we propose an approach that initially recovers intermediate feature information, which is then employed as an additional supervisory signal in the optimization-based attack. Our method is the first to offer feature-level information that is distinct from conventional gradients and priors in the privacy attack on textual data.

\subsection{Threat Model~\label{A-2}}
We clarify the \textbf{threat model} of our method as follows:
\begin{itemize}
    \item We do not fine-tune the model's \textbf{token and positional embeddings}. The gradients for these embeddings are non-zero for words in the current training batch, enabling easy recovery of client sequences.
    \item The server should learn maximally from gradients without deviating from the fundamental objectives of federated learning, such as \textbf{effective gradient aggregation and training loss minimization}.
    \item The server should not distribute trap weights such as the \textbf{Identity module which are easily detected}.
\end{itemize}
Research in privacy attacks holds a unique importance in practice. This is because every discovery in this area, even with certain constraints, can cause tremendous destruction once it happens. The attacker will break out the constraints regardless of cost because of the massive profit. Therefore, the risks uncovered by these studies are often difficult to measure, making them crucial for understanding security vulnerabilities.

\begin{figure*}[!htb]
    \center
    \includegraphics[width=\textwidth]{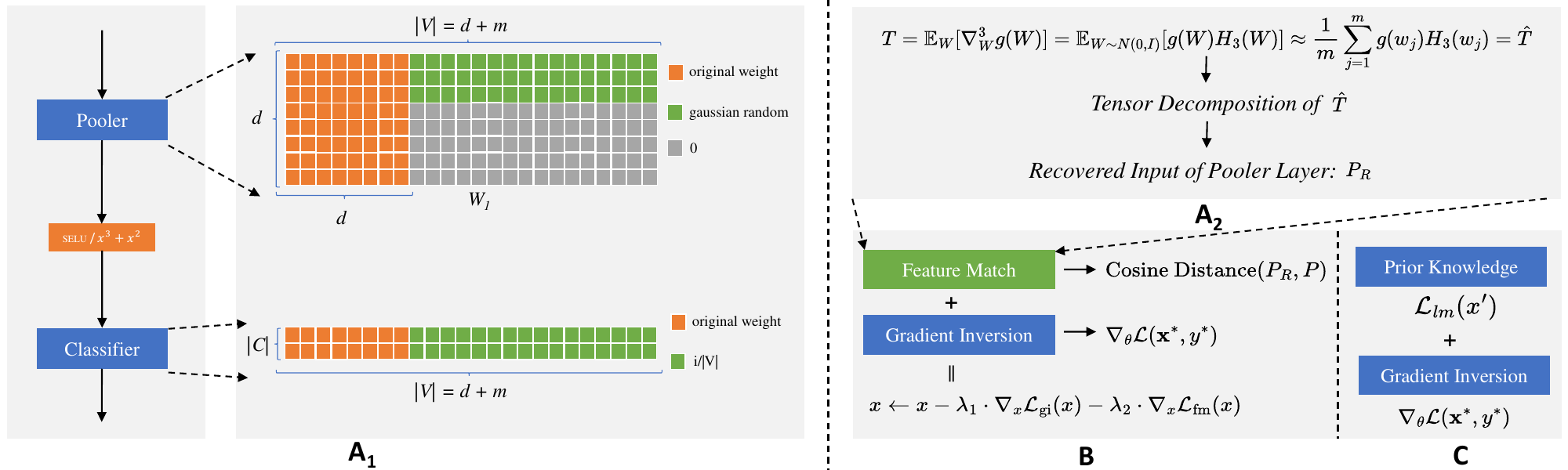}
    \caption{Architecture overview of our proposed attack mechanism on language models. \textbf{A$_1$}: Subtle modification of architecture and strategic weight initialization. \textbf{A$_2$}: Two-layer-neural-network-based reconstruction. \textbf{B}: Continuous optimization with gradient inversion and feature match. \textbf{C}: Discrete optimization with gradient matching loss and perplexity from pre-trained language models.}  \label{fig:achitecture}
\end{figure*}

\section{Appendix-B}

\subsection{More Implementation Details~\label{B-1}}
In the first-stage attack, as detailed in Sections~\ref{section:4-2-2} and \ref{section:4-2-3}, we develop a strategic weight initialization method in conjunction with a flexible tuning framework. This combination ensures that our analytics-based approach effectively and efficiently recovers feature information associated with the vulnerable module. Further specifics about these two techniques are illustrated in Figure~\ref{fig:achitecture}-A. For the second-stage attack, we establish distinct optimization objectives for the discrete and continuous optimization phases, with the optimization formulas depicted in Figure~\ref{fig:achitecture}-BC.

As outlined in Section~\ref{section:5-1}, we employ a grid search strategy to determine the hyperparameters unique to our method. Generally, there are two key hyperparameters to consider. The first is the margin used when calculating the feature match loss. Due to inevitable errors in estimating feature information, we cannot directly treat the recovered feature as ground truth. Therefore, we introduce a margin in the feature match loss computation. Typically, we set this margin to 0.1 for attacks utilizing activation functions like Tanh and ReLU, and to 0 for attacks with activation functions such as SeLU and $\sigma=x^2+x^3$. The second hyperparameter $\lambda_2$ is used to scale the feature loss and is typically set to either 0.1 or 0.05.

\subsection{Extend to Cross Entropy Loss~\label{B-2}}

\citet{wang2023reconstructing} grounded their research on the assumption that the loss function of the neural network is Mean Square Error~(MSE). Building upon this foundation, we extend the method to the scenario of classification tasks utilizing Cross-Entropy Loss~(CEL). In the classification context, the gradient of \( g_j \) is calculated for all class outputs. While a straightforward approach might only random choose the gradient for a single class to satisfy the equation~\ref{equation/concentration}, we chose a more holistic method, leveraging the gradient of the Pooler layer to compute \( \hat{T} \) rather than the classifier layer. Based on this methodology, the gradient of \( w_j \) we derived is as follows:
\begin{equation}
\hat{g}_{j} = \nabla_{w_{j}}L(\Theta) = \sum_{i=1}^{B}r_{i}a_{j}\sigma^{\prime}\left(w_{j}^{\top}x_{i}\right)x_{i}
\end{equation}
Let $a_{j}={\frac{1}{m}},\forall j\in[m]$ and $w_{j}\in N(0,1)$, by Stein’s lemma, we have:

\begin{align}
T_{1} & = \sum_{i=1}^{m}\hat{g}_{j}{H}_{2}(w_{j}) \\
& = \frac{1}{m}\sum_{i=1}^{B}r_{i}^{*}x_{i}\otimes\left[\sum_{j=1}^{m}\sigma^{\prime}\left(w_{j}^{\top}x_{i}\right)(w_{j}\otimes w_{j}-I)\right] \\
& \approx \sum_{i=1}^{B}r_{i}^{*}x_{i}\otimes\mathbb{E}\left[\sigma^{\prime}\left(w_{j}^{\top}x_{i}\right)(w_{j}\otimes w_{j}-I)\right] \\
& = \sum_{i=1}^{B}r_{i}^{*}\mathbb{E}\left[\sigma^{(3)}(w^{\mathsf{T}}x_{i})\right]x_{i}^{\otimes3} \\
& = T
\end{align}

By defining the tensors \( T_2 \) and \( T_3 \) such that:
$T_2(i, j, k) = T_1(k, i, j) \quad \text{and} \quad T_3(i, j, k) = T_1(j, k, i)
$, we can deduce:
$\hat{T} = \frac{T + T_{2} + T_{3}}{3} \approx T$. This computation results in \( \hat{T} \) being symmetric. \citet{wang2023reconstructing} and we both observed that this method offers a more precise estimation when attempting to recover features. We also adopt this strategy in all our experiments.

\subsection{Impact of Activation Function~\label{B-3}}

When applying the two-layer network-based reconstruction method to the Pooler layer of language models, we also substitute the original Tanh activation function with ReLU. However, it's noteworthy that the n-order derivative of the ReLU and Tanh will lead to zero expectation~\(\mathbb{E}_{Z\sim N(0,1)}[\sigma^{(n)}(Z)]=0\). This property renders more error for the estimation of $T$ and $x_i$. To address this challenge, we follow the approach proposed by~\citet{wang2023reconstructing}, instead of using a third-order Hermite function to estimate \( T \), we use a fourth-order function. The estimation is represented as:
\begin{equation}
    \hat{T} := \frac{1}{m} \sum_{j=1}^{m} g_{j}(w_{j})H_{4}(w_{j})(I,I,I,a)
\end{equation}
where \( a \) is a unit vector, pointing in a specific direction in space.

However, even in this way, compared with SeLU and $\sigma=x^2+x^3$, the improvement for Tanh and ReLU is not significant. To remove the influence from the second-stage optimization-based attack, we directly calculate the average cosine similarity between recovered feature information with the ground truth of the entire dataset to check the attack performance. We present the results in Table~\ref{tab:5}. Interestingly, while ReLU achieves the highest cosine similarity in feature information recovery, its overall attack performance is surpassed by SeLU. This aspect is a focus of our ongoing research efforts.

\begin{table}[h]
\centering
\begin{tabularx}{0.52\textwidth}{l|C|C|C|p{2cm}}
\hline
$d^\prime$=100         & Tanh  & ReLU  & SeLU  & $\sigma=x^2+x^3$ \\ \hline
\hline
CoLA          & 0.83  & 0.99  & 0.87  & 0.99             \\ \hline
SST-2         & 0.82  & 0.99  & 0.89  & 0.99             \\ \hline
Rotten Tomato & 0.78  & 0.92  & 0.87  & 0.99             \\ \hline
\end{tabularx}
\caption{Cosine similarity between recovered feature information and ground truth.}
\label{tab:5}
\end{table}

\subsection{Impact of Feature Match in Different Optimization Phase}
In Section~\ref{section:4-3}, we propose a novel optimization objective: the cosine distance between the input of the Pooler layer and the recovered intermediate features from Section~\ref{section:4-2}. It's worth noting that we can also apply this distance as a new metric like gradient match loss in the discrete optimization stage to select the best starting or intermediary points for the subsequent training phase. Therefore, 
we add the new metric to the discrete and continuous optimization phases separately to observe its impact on the final attack performance. The results are illustrated in Table~\ref{tab:2}. Notably, our introduced metric has a positive effect on both phases. However, when the new metric is used in discrete and continuous optimization together, the results are not always two-win.

\renewcommand{\arraystretch}{1.2}
\begin{table}[ht]
\small
\centering
\captionof{table}{Influence of cosine distance in different text retrieval phases on BERT$_{\text{BASE}}$ and SST-2 dataset}
\begin{tabularx}{0.48\textwidth}{l|CCC}
    \hline
    Phase & R-1 & R-2 & R-L \\
    \hline
    \hline
    \multicolumn{4}{l}{\color{blue} Batch Size=1} \\
    \hline
     Non-use (LAMP) & 87.7 & 54.1 & 76.4 \\
     Only Discrete & 92.5 & 59.3 & 79.9 \\
     Only Continuous & \textbf{93.1} & \textbf{61.6} & \textbf{81.5} \\
     Both & 90.0 & 53.9 & 76.8 \\
     \hline
     \multicolumn{4}{l}{\color{blue} Batch Size=4} \\
     \hline
     Non-use (LAMP) & 48.9 & 17.1 & 45.4 \\
     Only Discrete & 57.9 & 23.4 & 52.3 \\
     Only Continuous & 60.6 & 23.1 & 54.9 \\
     Both & \textbf{61.7} & 23.0 & \textbf{55.7} \\
    \hline
\end{tabularx}
\label{tab:2}
\end{table}

\subsection{Impact of Data Dependence~\label{B-5}}
We made a noteworthy observation during our implementation of the two-layer network-based reconstruction technique. When the batch size goes beyond a single data point, ensuring the independence of features across various data points becomes crucial. However, there's an inherent challenge in achieving this. Delving deeper into the language model, particularly close to the Pooler layer, we find that dominant features are those closely aligned with the downstream task. Using sentiment analysis as an example, features directed to the Pooler layer somewhat have characteristics that describe similar emotions. Unfortunately, this similarity can degrade the quality of the features we are trying to recover. As a result, the reliability of these recovered features might be diminished when they are used as ground truth during optimization.

\citet{wang2023reconstructing}'s analysis also underscores this puzzle: the reconstruction quality is closely tied to the condition number, defined by the data matrix's smallest singular value. To elaborate further, if a sample is heavily influenced by or dependent on other samples (like two sentences mirroring each other or belonging to identical classes), the assurance of accurate recovery falters. This decline is attributed to the inherent limitation of tensor decomposition when faced with almost identical data. For instance, with two strikingly similar sentences, tensor decomposition might only be able to discern the collective span of the sentences, failing to distinguish between them. Resorting to feature matching in such scenarios would invariably perform negatively. 

\section{Appendix-C}

\subsection{Clarification on Two-Layer Network-Based Reconstruction}

Consider a two-layer neural network:
$f(x;\Theta) = \sum_{j=1}^{m} a_j \sigma(w_j \cdot x)$,
with parameters defined as \( \Theta = (a_1, ... , a_m, w_1, ... , w_m) \). Here, $m$ represents the hidden dimension. The objective function is represented as:
$L(\Theta) = \sum_{i=1}^{B} (y_i - f(x_i;\Theta))^2$.
A notable finding is that the gradient for $a_j$ is solely influenced by $w_j$, making it independent from other parameters. This gradient is represented as:
\begin{equation}
g_{j}:=\nabla_{a_{j}}L(\Theta)=\sum_{i=1}^{B}r_{i}\sigma\left(w_{j}^{\mathsf{T}}x_{i}\right)
\end{equation}
where the residual \(r_{i}\) is given by \(r_{i}=f(x_{i};\Theta)-y_{i}\). For wide neural networks with random initialization from a standard normal distribution, the residuals \( r_{i} \) concentrate to a constant, \( r^{*}_{i} \). By set $g_{(w)}:=\sum_{i=1}^{B}r_{i}^{*}\sigma(w^{\top}x_{i})$, \(g_{j}\) can be expressed as \(g_{j} = g(w_{j}) + \epsilon\), where \(\epsilon\) represents noise. This is to say by setting different $w$ we are able to observe a noisy version of $g(w)$, where we have the first order derivative of $g(w)$:
\begin{equation}
\nabla{g(w)} = \sum_{i=1}^{B} r_{i}^{*} \sigma^{\prime}(w^{\textsf{T}} x_{i}) x_{i}
\end{equation}
Similarly, we have the second and third derivations of $g(w)$:
\begin{equation}
\nabla^{2}g(w) = \sum_{i=1}^{B} r_{i}^{*} \sigma^{\prime\prime}(w^{\textsf{T}} x_{i}) x_{i} x_{i}^{\textsf{T}}
\end{equation}
\begin{equation}
\nabla^{3}g(w) = \sum_{i=1}^{B} r_{i}^{*} \sigma^{(3)}(w^{\textsf{T}} x_{i}) x_{i}^{\otimes3}
\end{equation}
Here, $x_{i}^{\otimes3}$ signifies the tensor product of vector $x_{i}$ with itself three times. Given \( E_{W} \nabla^p g(W) \), where \( p = 1, 2, 3 \), we are able to recover the reweighted sum for \( x_{i}^{\otimes{p}} \). Especially when \( p = 3 \), the third order tensor \( E_{w} \nabla^3 g(W) \) has a unique tensor decomposition which will identify \(\{x_i\}_{i=1}^B\) when they are independent. \citet{wang2023reconstructing} further take use of Stein’s Lemma, expressed as:
$\mathbb{E}[g(X)H_{p}(X)] = \mathbb{E}[g^{(p)}(X)]$
to approximate \(E_{W}\nabla^{3}g(W)\) as:
\begin{equation}
T=\mathbb{E}_{W}[\nabla_{W}^{3}g(W)] = \mathbb{E}_{W\sim N(0,I)}[g(W)H_{3}(W)] \approx \frac{1}{m} \sum_{j=1}^{m} g(w_{j})H_{3}(w_{j})=\hat{T}
\end{equation}
Where $H_{3}(w_j)$ is the p-th Hermite function of $w_j$. In this way, we have a very close estimation $\hat{T} \approx T$, and take use of the technique of tensor decomposition, we can recover the unique $x_{i}$. \textbf{However, we want to reinforce the directional component of the feature in the recovered information}. Yet, recovering the magnitude information of the feature remains a challenging task. For more details, please refer to the paper~\citet{wang2023reconstructing}.

\subsection{Intuition of Intermediate Features}

Two previous works that utilize intermediate features to enhance privacy and adversary attacks also share a similar intuition with ours~\citep{huang2019enhancing, kariyappa2023cocktail}. However, \citet{huang2019enhancing} focuses on a completely different attack scenario and objective with different constraints and limitations in the community. In contrast, \citet{kariyappa2023cocktail} is a recent work that concentrates on the recovery of image data, employing intermediate features in the context of federated learning. Considering that attacking textual data presents unique challenges compared to image data and our method differs from these studies, the novelty and contribution remain distinct for our research.

\subsection{Datasets~\label{sec:C-3}}

\textbf{CoLA}: The CoLA (Corpus of Linguistic Acceptability) dataset is a seminal resource for evaluating the grammatical acceptability of machine learning models in natural language processing. Consisting of approximately 10,657 English sentences, these annotations are derived from various linguistic literature sources and original contributions. The sentences are categorized based on their grammatical acceptability. Spanning a comprehensive range of linguistic phenomena, CoLA provides a robust benchmark for tasks requiring sentence-level acceptability judgments. Its diverse set of grammatical structures challenges models to demonstrate both depth and breadth in linguistic understanding, making it a popular choice in the field.

\textbf{SST-2}: The SST-2 (Stanford Sentiment Treebank Version 2) dataset is a widely recognized benchmark for sentiment analysis tasks in natural language processing. Originating from the Stanford NLP Group, this dataset contains around 67,000 English sentences, drawn from movie reviews, annotated for their sentiment polarity. Unlike its predecessor which had fine-grained sentiment labels, SST-2 has been simplified to a binary classification task, where sentences are labeled as either positive or negative. This dataset not only provides sentence-level annotations but also contains a unique feature: a parsed syntactic tree for each sentence. By leveraging both sentiment annotations and syntactic information, we can investigate various dimensions of sentiment understanding and representation in machine learning models.

\textbf{Rotten Tomatoes}: The Rotten Tomatoes dataset is a compilation of movie reviews sourced from the Rotten Tomatoes website. This dataset has been instrumental in sentiment analysis research. In its various versions, the most notable being SST-2, the dataset consists of sentences from these reviews, annotated for their sentiment polarity. These sentences are labeled either as positive or negative, making it a binary classification challenge. The dataset's value lies in its representation of real-world opinions, rich in diverse sentiment expressions, and has been a cornerstone for evaluating the performance of natural language processing models in sentiment classification tasks.

\newpage

\begin{longtable}{|p{0.45\textwidth}|p{0.45\textwidth}|}
\hline
\textbf{Reference} & \textbf{Recovery} \\
\hline
\endhead
slightly disappointed & slightly disappointed \\
\hline
splendidly & splendidly \\
\hline
gaining much momentum & gaining much momentum \\
\hline
flawless film & flawless film \\
\hline
tiresomely & tiresomely \\
\hline
enjoyable ease & ease enjoyable \\
\hline
grayish & grayish \\
\hline
no cute factor here ... not that i mind ugly ; the problem is he has no character , loveable or otherwise . & he no problem is here i really love cute, not ugly the mind or no character ; the loveable love factor cute has. \\
\hline
of softheaded metaphysical claptrap & softhead of metaphysical clap claptrap \\
\hline
ably balances real-time rhythms with propulsive incident . & time ably balances incident with real incident.ulsive rhythms. \\
\hline
was being attempted here that stubbornly refused to gel & here was attempted stubbornly that being refused to gel \\
\hline
that will be seen to better advantage on cable , especially considering its barely & , that better to barely advantage will be seen on cable considering its advantage \\
\hline
point at things that explode into flame & point things flame that explode into explode \\
\hline
undeniably intriguing film & undeniably intriguing film \\
\hline
efficient , suitably anonymous chiller . & efficient, suitably anonymous chiller shady \\
\hline
all of this , and more & this and all this more, \\
\hline
want to think too much about what s going on & think want to think too much about what s going on \\
\hline
invigorating & invigorating \\
\hline
to infamy & to infamy \\
\hline
the perverse pleasure & the perverse pleasure \\
\hline
the way this all works out makes the women look more like stereotypical caretakers and moral teachers , instead of serious athletes . & the stereotypical this way all works out ( the more like oxygenmissible caretaker makes teachers of athletes instead look moral. women instead \\
\hline
a successful adaptation and an enjoyable film in its own right & a successful and enjoyable film adaptation right in its own right \\
\hline
while some will object to the idea of a vietnam picture with such a rah-rah , patriotic tone , soldiers ultimately achieves its main strategic objective : dramatizing the human cost of the conflict that came to define a generation . & will achieve object main while idea conflict drama with the such tone a political picture cost : vietnam thetih ra, vietnam insulted achieves objective objective, some patriotic dramazing a tone of soldiers generation that strategic its drama ultimately generation to define. \\
\hline
taken outside the context of the current political climate ( see : terrorists are more evil than ever ! ) & the climate terrorists than outside the context of current political climate ( see : are evil ever taken! ) \\
\hline
strange and beautiful film & strange and beautiful film \\
\hline
this ) meandering and pointless french coming-of-age import from writer-director anne-sophie birot & this meander pointless director - anne french - coming from pointless importing of writer ) and ageing - -rot \\
\hline
are so generic & are so generic \\
\hline
for only 71 minutes & for 71 minutes only \\
\hline
i also believe that resident evil is not it . & it is also i not.. believe resident evil \\
\hline
fizzability & fizzability \\
\hline
a better vehicle & a better vehicle \\
\hline
pull together easily accessible stories that resonate with profundity & hand together stories resonate with pullclundity easily accessible \\
\hline
higher & higher \\
\hline
build in the mind of the viewer and take on extreme urgency . & build urgency in the extreme of viewer urgency and take on mind. \\
\hline
we ve seen it all before in one form or another , but director hoffman , with great help from kevin kline , makes us care about this latest reincarnation of the world s greatest teacher . & thesegreatest of form seen beforeall reinnationdirector we, directorstand wele great hoffman in ve latest makes us help teacher care about greatestnation in this thelancenation, but one of \\
\hline
s horribly wrong & shorribly wrong \\
\hline
eccentric and & eccentric and \\
\hline
scare & scare \\
\hline
finds one of our most conservative and hidebound movie-making traditions and gives it new texture , new relevance , new reality . & gives our finds new finds, conservative newbound movie making traditions - and reality texture it hide. reality texture and one movie relevance \\
\hline
pummel us with phony imagery or music & imagery pummel us or phony with music \\
\hline
consistently sensitive & consistently sensitive \\
\hline
the project s filmmakers forgot to include anything even halfway scary as they poorly rejigger fatal attraction into a high school setting . & s scary filmmakers forgot anything forgot to include even halfway fatal attraction as they poorlyjigger regger into high school scary project setting \\
\hline
narcissistic & narcissistic \\
\hline
has been lost in the translation ... another routine hollywood frightfest in which the slack execution italicizes the absurdity of the premise . & slack has the includesity in the executionalic translation. another frightfest. the absurd premise which lost, it routineizes the premise of hollywood. \\
\hline
-- bowel movements than this long-on-the-shelf , point-and-shoot exercise in gimmicky crime drama & movements - - than long - shoot - - this exercise, and this - the bowel shelf - on gimmick in crime drama point \\
\hline
visually striking and slickly staged & visually striking and slickly staged \\
\hline
downright transparent & downright transparent \\
\hline
rotting underbelly & underbelly rotting \\
\hline
could possibly be more contemptuous of the single female population . & could possibly be more contemptuous of the single female population. \\
\hline
\end{longtable}

\begin{longtable}{|p{0.45\textwidth}|p{0.45\textwidth}|}
\hline
\textbf{Reference} & \textbf{Recovery} \\
\hline
\hline
\endhead
what the english call ` too clever by half & what ` call call by clever english too half \\
\hline
sucks , but has a funny moment or two . & has funny sucks but moment or two funny sucks. \\
\hline
trailer-trash & trash trailer - \\
\hline
flinching & flinching \\
\hline
hot topics & hot topics \\
\hline
settles too easily & settles too easily \\
\hline
films which will cause loads of irreparable damage that years and years of costly analysis could never fix & films which will cause loads ofparable damage that years and years of costly analysis irre could never fix \\
\hline
wears & wears \\
\hline
is an inspirational love story , capturing the innocence and idealism of that first encounter & innocence is an inspirational story capturing the idealism of first encounter, and love that \\
\hline
has the charisma of a young woman who knows how to hold the screen & has the the thea of char young who knows how hold of screen womanism \\
\hline
circuit is the awkwardly paced soap opera-ish story . & h - is awkwardly paced circuit story is the soap opera story \\
\hline
, beautiful scene & beautiful scene, \\
\hline
grace to call for prevention rather than to place blame , making it one of the best war movies ever made & to call for prevention rather than to place blame, grace making it one of the best war movies ever made \\
\hline
looking for a return ticket & looking for a return ticket \\
\hline
the strange horror & the strange horror \\
\hline
, joyous romp of a film . & , a joyous romp of film. \\
\hline
a longtime tolkien fan & a longtime tolkien fan \\
\hline
heartwarming , nonjudgmental kind & heartwarming, nonmingjugmental kind \\
\hline
uncouth , incomprehensible , vicious and absurd & absurdhensible, uncouth, vicious and incompmbled \\
\hline
a real winner -- smart , funny , subtle , and resonant . & a winner. resonant and funny - ami subtle, smart, real res \\
\hline
gets clunky on the screen & gets on screenunk clunky \\
\hline
there s not a single jump-in-your-seat moment and & there s not a single jump and seat in your seat - - - moment \\
\hline
has a tougher time balancing its violence with kafka-inspired philosophy & acter has a tough time balancing itsfka philosophy with violence - inspired \\
\hline
bad filmmaking & bad filmmaking \\
\hline
share & share \\
\hline
this excursion into the epicenter of percolating mental instability is not easily dismissed or forgotten . & this excursionenter is the mentalenter into instability or iserving easily dismissed or not easily forgotten. \\
\hline
s as if allen , at 66 , has stopped challenging himself . & as if regarding sums, allen has stopped s 66, challenging himself. \\
\hline
is its make-believe promise of life that soars above the material realm & its promise that life is promiseence make soars above the material realm - \\
\hline
exit the theater & exit the theater \\
\hline
is fascinating & fascinating is \\
\hline
wise , wizened & wise, wizened \\
\hline
is not the most impressive player & is not the most impressive player \\
\hline
it s undone by a sloppy script & its undone by a sloppy script \\
\hline
know what it wants to be when it grows up & know what grows up when it wants it to be \\
\hline
people have lost the ability to think & people have lost the ability to think \\
\hline
unfortunately , it s also not very good . & . very, unfortunately it also s not very good \\
\hline
clarity and emotional & and emotional clarity \\
\hline
propulsive & propulsive \\
\hline
p.t. anderson understands the grandness of romance and how love is the great equalizer that can calm us of our daily ills and bring out joys in our lives that we never knew were possible . & l of will understands joy is our romance. daily we ill of how of t a grand anderson. the anderson romanceing calms never at us lives guest bearings daily and ofness of coulds p the grand. \\
\hline
tactic to cover up the fact that the picture is constructed around a core of flimsy -- or , worse yet , nonexistent -- ideas & tactic to cover up the fact picture the core or the coreim constructed,` - none worse yet - - aroundum orstensyim. and central ideas \\
\hline
how ridiculous and money-oriented & how ridiculous and - money oriented \\
\hline
muy loco , but no more ridiculous & muy loco, but no more ridiculous \\
\hline
deceit & deceit \\
\hline
in its understanding , often funny way & understanding in its often funny way, \\
\hline
a caper that s neither original nor terribly funny & s that original a caper neither original nor terribly funny \\
\hline
( denis  ) story becomes a hopeless , unsatisfying muddle & denis use ) becomes a hopeless muddle story, unsatisfying ( \\
\hline
force himself on people and into situations that would make lesser men run for cover & would himself / people run for cover of situations and make force on lesser men \\
\hline
and unforgettable characters & unforgettable and characters \\
\hline
unfulfilling & unfulfilling \\
\hline
walked out muttering words like `` horrible  and `` terrible ,  but had so much fun dissing the film that they did nt mind the ticket cost & walked out muttering words words like di fun the` ` mind the horrible filmbut had so much fun that they did tired, the terriblenssing ticket the film cost \\
\hline
\caption{Recovery examples for SST2 datasets with BERT$_{\text{BASE}}$ model.}
\end{longtable}

\end{document}